\pdfoutput=1

\documentclass[11pt]{article}

\usepackage{EMNLP2023}
\usepackage{times}
\usepackage{latexsym}
\usepackage{enumitem}
\usepackage[T1]{fontenc}
\usepackage[utf8]{inputenc}
\usepackage{graphicx}
\usepackage{booktabs}
\usepackage{multirow}
\usepackage{microtype}
\usepackage{url}
\usepackage{inconsolata}

\title{Descriptive Knowledge Graph in Biomedical Domain}

\author{Kerui Zhu \quad Jie Huang$^\dag$ \quad Kevin Chen-Chuan Chang \\
University of Illinois at Urbana-Champaign, USA \\ 
\texttt{\{keruiz2, jeffhj, kcchang\}@illinois.edu}}

\begin{document}
\maketitle
\begin{abstract}

We present a novel system that automatically extracts and generates informative and descriptive sentences from the biomedical corpus and facilitates the efficient search for relational knowledge. Unlike previous search engines or exploration systems that retrieve unconnected passages, our system organizes descriptive sentences as a relational graph, enabling researchers to explore closely related biomedical entities (e.g., diseases treated by a chemical) or indirectly connected entities (e.g., potential drugs for treating a disease). Our system also uses ChatGPT and a fine-tuned relation synthesis model to generate concise and reliable descriptive sentences from retrieved information, reducing the need for extensive human reading effort. With our system, researchers can easily obtain both high-level knowledge and detailed references and interactively steer to the information of interest. We spotlight the application of our system in COVID-19 research, illustrating its utility in areas such as drug repurposing and literature curation.\footnote{Demo video: \url{https://www.youtube.com/watch?v=VvWs9JEP8ro} \quad \quad System website: \url{https://zhukerui.github.io/CovidDEER/}. $^\dag$Corresponding author.}

\end{abstract}

\section{Introduction}

Efficiently extracting knowledge from the vast and ever-growing corpus of literature is crucial for researchers to keep up with the latest discoveries and trends in the field. The COVID-19 pandemic has highlighted this need, with thousands of related studies being published in a short period when a new disease emerges. However, surveying the latest findings requires significant effort, and researchers may struggle to see the big picture, leading to duplicated work and delaying the development of treatments \cite{wang-etal-2021-covid}. Hence, an exploration system that can effectively retrieve comprehensive information from the latest literature corpus is important.

Existing exploration systems manage information in generally three granularities: documents, sentences, and knowledge facts. Document-level retrieval usually takes keyphrases \cite{Shen2018EntityApproach} or questions \cite{10.1145/3451964.3451965, Levy2021Open-DomainDomains} as queries and finds relevant documents. Using such systems, researchers need to read the retrieved documents to find relevant information, which is still time-consuming \cite{wang-etal-2020-evidenceminer}. Sentence-level \cite{Wang2021TextCOVID-19, wang-etal-2020-evidenceminer, Lahav2022ADirections} retrieval usually takes entities, entity types, or sentences as queries and finds sentences that contain the entities and entity types or are semantically similar to the input sentence. The retrieved sentences require less reading effort, but they are retrieved as independent text pieces, which don't provide a general overview of the knowledge. Knowledge facts, on the other hand, are usually (head, relation, tail) triples extracted from the corpus and stored as a knowledge graph, which concisely reveals the connection between entities. However, systems that retrieve knowledge facts \cite{ChungBioKDE:Platform, wang-etal-2021-covid} usually allow queries for entities and entity types only, and the retrieved knowledge facts can only cover the relations in a fixed pre-defined set.

\begin{figure*}[t]
    \centering
    \includegraphics[width = \textwidth]{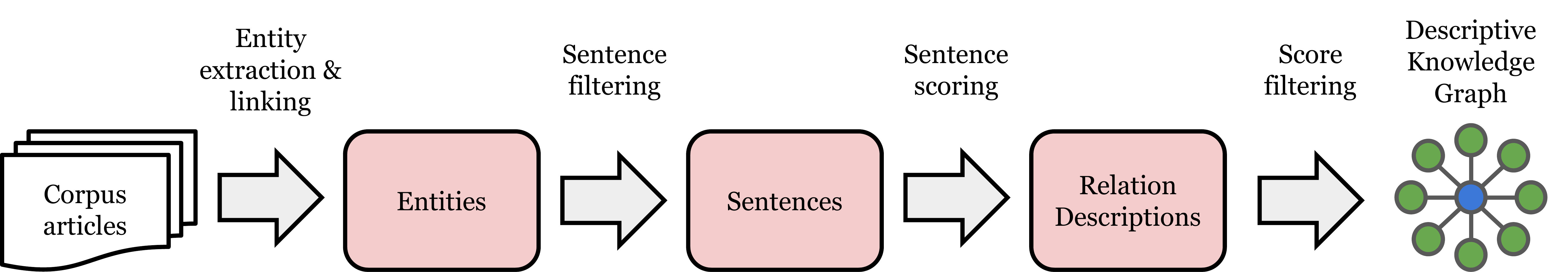}
    \caption{Data pipeline for descriptive knowledge graph construction: 1. Extract entities from corpus articles. 2. Remove sentences with missing subject or object entities. 3. Score sentences as relation descriptions with the RDS. 4. Filter low-score relation descriptions (score < 0.7) and build the graph.}
    \label{fig:pipeline}
    \vspace{-2mm}
\end{figure*}

To overcome the limitations mentioned above, we develop an exploration system that manages corpus sentences as a \textit{descriptive knowledge graph} \cite{huang-etal-2022-deer}. The Descriptive knowledge graph for Explaining Entity Relationships (DEER) is a special knowledge graph where each edge is not a relation label but a set of relational sentences describing the relationship between a pair of entities \cite{handler2018relational,huang2022open,huang2022ver,liu-etal-2023-dimongen,huang2023ccgen}. We collect entities and sentences from the biomedical corpus to build a domain-specific DEER and provide useful tools for users to effectively query and explore the graph.

Our system allows users with little prior knowledge to interactively retrieve up-to-date, comprehensive, and easily understandable relational sentences, and explore relationships between entities in one-hop or multi-hop connections. Additionally, we use ChatGPT and a fine-tuned relation synthesis model to generate succinct relation descriptions for entity pairs based on the retrieved sentences to aid users' reading. It is worth mentioning that our system is automatically built without any supervised training or hand-crafted rules, making it seamlessly adaptable to any biomedical corpus with ease, and it can serve as a frontrunner for collecting knowledge in any future emergency.

\section{Graph Construction}\label{sec:construction}

DEER \cite{huang-etal-2022-deer} is a form of knowledge representation that balances the openness and informativeness of free text and the structured representation of the knowledge graph. In this graph, nodes are entities, and edges are sentences describing the relationship between the two nodes, called \textit{relation descriptions}, pointing from the subject node to the object node in the sentences.  
The previous DEER graph \cite{huang-etal-2022-deer} was built upon Wikipedia. Due to the limitations of the corpus, it does not contain much biomedical domain knowledge. In this section, we will introduce techniques for building a descriptive knowledge graph in the biomedical domain.
Based on that, users could retrieve sentences with efficient graph queries and view the result from a connected perspective to gain a more holistic understanding of the retrieved information.

\subsection{Corpus}

To efficiently establish the system for retrieving knowledge about a specific topic, we build the DEER on a sub-domain corpus. In this work, we use COVID-19 Open Research Dataset (CORD-19) \cite{wang-etal-2020-cord} as a representative corpus in the biomedical domain, which comprises scientific papers related to COVID-19 and other coronaviruses, and note the DEER built from it as \textit{CovidDEER}. For demonstration purposes, we used the snapshot on August 8th, 2020 to simulate a corpus when a new disease outbreak and some clinical experimental results have been published. With this corpus, we demonstrate how our system can retrieve valuable information for disease research and drug repurposing.

\begin{table}[tp!]
\begin{center}
\begin{tabular}{l|l|l}
\toprule
\textbf{\# documents} & \textbf{\# nodes} & \textbf{\# edges}\\
\midrule 
72,014 & 140,574 & 863,102 \\ 
\bottomrule
\end{tabular}
\vspace{-1mm}
\caption{Corpus and graph statistics for CovidDEER, collected using the RDS threshold of 0.7.}
\vspace{-3mm}
\label{table:corpus_stat}
\end{center}
\end{table}

\subsection{Pipeline}

To construct \textit{CovidDEER}, our system employs a pipeline that processes the corpus as follows:

\paragraph{Entity Extraction and Linking}\label{linking} Initially, we extract biomedical entities from each sentence in the corpus and link them to biomedical ontologies using the NCBI Pubtator API and the SciSpacy library \cite{Neumann2019ScispaCy:Processing}. Specifically, we link the extracted entities to Cellosaurus, OMIM, MeSH, Gene, Taxonomy, and UMLS metathesaurus.

\paragraph{Sentence Filtering} Next, we use SciSpacy to parse the sentences and remove those which do not have a subject entity or object entity as these entities serve as the head and tail entities in the relation description. Missing head or tail entities indicate that these sentences are not appropriate for describing relationships.

\paragraph{Sentence Scoring} Then, we gather the parameters for a scoring function and use it to score the sentences. We use the relation description score (RDS) introduced in \citet{huang-etal-2022-deer} as the scoring function. This scoring function extracts the dependency path between the head entity, tail entity and other relation-related words in a sentence and generates a score between 0 and 1 to indicate how well this sentence expresses the relationship of the entities. Higher score indicates better the sentence as a relation description. A domain-specific RDS scoring function requires data of dependency path frequency from the domain corpus. Once the scoring function is setup with adequate corpus data, it can be frozen to evaluate any in-domain sentences. In this work, we collect dependency path frequencies from the whole CORD-19 corpus.

For each pair of subject-object entities in a sentence, we apply the scoring function and store the sentence and the RDS score with the corresponding entity pair.

\paragraph{Score Filtering} For each entity pair, we filter out the low-quality sentences with the RDS score and assign the rest to the edge from the head entity to the tail entity to construct the DEER. In practice, a sentence with RDS score over 0.7 usually has a good quality.

A visualization of the data pipeline is depicted in Figure~\ref{fig:pipeline} and the statistics of the CORD-19 corpus and \textit{CovidDEER} are listed in Table \ref{table:corpus_stat}. Note that the graph can be easily updated with the latest knowledge by extracting relation descriptions from recent papers.

\begin{figure*}[t]
    \centering
    \includegraphics[width = 0.9\textwidth]{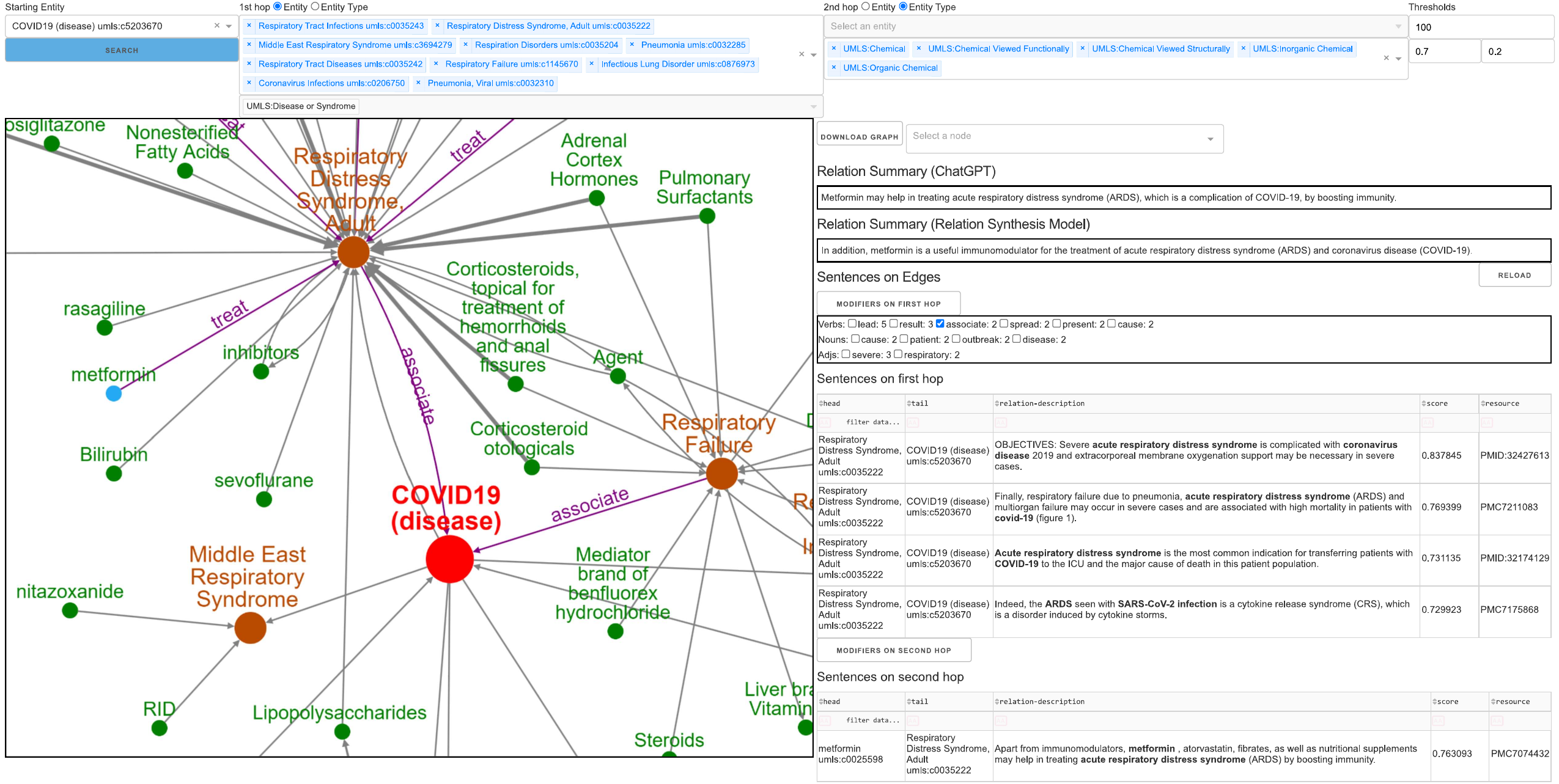}
    \caption{The web interface of \textit{CovidDEER}. The interface shows a graph retrieved by a two-hop query: \textit{``COVID19''} -- 10 \textit{``Disease or Syndrome''} entities -- 5 \textit{``Pharmacologic Substance''} related entity types. The \textit{metformin} is selected (in blue) and a directed path, \textit{COVID19} $\rightarrow$ \textit{Respiratory Distress Syndrome, Adult} $\rightarrow$ \textit{metformin}, is used for relation summary by ChatGPT and relation synthesis model.}
    \label{fig:illustration}
\end{figure*}

\section{Graph Query}\label{sec:graph_query}

To retrieve sentences from the \textit{CovidDEER}, our system provides a Graph Query module with some auxiliary tools to allow an interactive and flexible search. Below are the queries supported by the graph query module.

\paragraph{Entity-Entity Query}

Entity-Entity query allows users to retrieve relation descriptions between two entities. This is achieved by extracting sentences that lie on the edge connecting the two entities in the constructed CovidDEER graph. Unlike from systems that retrieve all or random sentences where the entities co-occur, our system focuses on returning sentences that capture the primary relationships between the target entities. This gives users more informative and clearer sentences and saves users from being distracted by meaningless sentences.

\paragraph{Entity-Type Query}

To obtain a more comprehensive overview of relationships between an entity and an entity type, our system also supports Entity-Type query, which will retrieve sentences from edges between an entity and all its neighbors belonging to an entity type. For instance, users can set the entity to \textit{COVID-19} and the entity type to \textit{Chemicals}. Then the system will return relation descriptions between COVID-19 and all kinds of related chemicals, which provides some insights into the Chemical-Disease interactions related to COVID-19. Our system supports all the entity types in the ontologies mentioned in Section \ref{linking}.

\begin{table}[tp!]
\small
\setlength\tabcolsep{1.5pt}
\begin{center}
\begin{tabular}{l|p{0.85\linewidth}}
\toprule
\textbf{Type} & \textbf{Frequent Modifiers} \\
\midrule 
Nouns & treatment (14), chloroquine (6), efficacy (4), hydroxychloroquine (4), therapy (2), option (2), patient (2) \\
\hline
Verbs & show (7), treat (5), use (5), propose (2), include (2) \\
\hline
Adjs & apparent (4), antiviral (3), effective (3), safe (2), severe (2), antimalarial (2) \\
\bottomrule
\end{tabular}
\caption{Frequent Modifiers between Chemicals and COVID-19.}
\vspace{-4mm}
\label{table:frequent_modifiers}
\end{center}
\end{table}

\paragraph{Multi-hop Query}

Besides finding direct neighbors of an entity, users can also query multi-hop neighbors to explore more indirect connections. By specifying the entities or entity types at each hop, users can retrieve sentences for multi-hop inference. For example, a user may begin with \textit{COVID-19}, set \textit{Symptom} as the first-hop entity type and \textit{Chemical} as the second-hop entity type to explore drugs that can treat COVID-19 related symptoms, and thus, could be used for COVID-19 treatment. With this tool, our system could beat traditional knowledge graphs by providing the contextualized knowledge, and beat the text-based search engines by allowing multi-hop retrieval with one query.

\paragraph{Modifier Filtering}

When querying a popular entity, \textit{CovidDEER} may return too many edges, which may distract users from catching the general relationships. 
To alleviate this,
we define the words in a sentence that convey the relation information as the \textit{modifiers} and allow users to locate interesting edges using the modifiers. We extract the noun phrases, verbs, and adjectives on the dependency path between the two entities as the modifiers. For example, Table \ref{table:frequent_modifiers} shows the frequent modifiers collected between \textit{COVID-19} and its \textit{Chemical} neighbors. These modifiers provide insights into the \textit{COVID-19}-\textit{Chemical} relationships and users can click the modifiers to highlight the edges where they occur. This tool could also help users perform a more fine-grained query to reduce unwanted results.

Figure \ref{fig:illustration} shows an example interface of our system, where the retrieved results are displayed as a graph, and users can checkout sentences by clicking the edges.

\section{Relation Synthesis Model}\label{sec:relation_synthesis_model}

Although \textit{CovidDEER} displays the relational sentences in a graph view to reveal the connections between entities, considerable manual effort is still required to read and digest the information on the edges. Performing multi-hop logic inference is even harder as users need to find associated sentences across different edges. To reduce users' reading effort, we trained a relation synthesis model \cite{huang-etal-2022-deer}, which is based on a Fusion-in-Decoder model \cite{Izacard2020LeveragingAnswering} trained to take sequences of relation descriptions from the multi-hop paths between two entities in DEER and generate one single relation description for the entities. Each training data is collected by selecting the highest RDS-scored sentences on each edge in the multi-hop paths between a target entity pair as the input and the highest RDS-scored sentence on the one-hop path between the target pair as the output. In order to allow summarizing relation descriptions on each edge, we also add the lower-scored sentences on the one-hop path into the input. 
Since large language models have demonstrated strong capabilities through simple prompting \citep{openai2023gpt4,anil2023palm,qin2023chatgpt,bubeck2023sparks,huang-chang-2023-towards}, in addition to the fine-tuned model, we also prompt ChatGPT~\cite{openai2022chatgpt} to generate a short passage to summarize the relationship from the retrieved sentences. Detailed steps for fine-tuning and the prompt for ChatGPT can be found in Appendix \ref{sec:fine-tune} \& \ref{sec:prompt}. By reading the generated relation descriptions first, users can get a general idea of the relation between the entities and then decide whether to read the retrieved sentences or not.

\section{System Demonstration \& Evaluation}\label{sec:evaluation}

In this section, we first evaluate the relation synthesis model by assessing the faithfulness of the generation with respect to the input relation descriptions. Then we demonstrate our system's capacity in discovering unknown knowledge and locating information of interest with a drug repurposing task and a literature curation task respectively.

\subsection{Relation Synthesis Model Evaluation}

\citet{huang-etal-2022-deer} have demonstrated the capability of the relation synthesis model to generate easily understandable relation descriptions. However, in the biomedical domain, it is crucial for the model to generate truthful sentences and not mislead the reader with erroneous information. Table \ref{table:extract_generated} provides an example of the model's generation, where the extracted relation descriptions for (\textit{COVID-19}, \textit{Pneumonia}) and (\textit{Pneumonia}, \textit{Vaccines}) are the inputs to the model. The 1-hop relation summary is the summarized relation description over the sentences of one pair of entities, and the 2-hop relation summary is the synthesized relation description for (\textit{COVID-19}, \textit{Vaccines}) through aggregating the 2-hop path (\textit{COVID-19}, \textit{Pneumonia}, \textit{Vaccines}).

To evaluate the model's faithfulness, the authors of this work used the model to generate relation descriptions for 20 randomly selected samples from the test dataset, tried to find supporting evidence from the input and gave a score from 1 to 5 for each generation to indicate its faithfulness to the input. 
The final average score for the 20 samples is 4.10, indicating that the generation is generally supported by the input. However, there is still a gap before we can fully trust it and we suggest users read the retrieved sentences to acquire reliable knowledge and only use the generated relation description as a reference.

\subsection{Case Study 1: Drug Repurposing}

Drug repurposing intends to identify new uses for drugs that were originally used to treat other diseases. \textit{CovidDEER} can aid researchers in identifying candidate drugs through the following steps:
\begin{itemize}[noitemsep,nolistsep,leftmargin=*]
\item Set the target disease as the starting node.
\item Search the first-hop neighborhood for diseases and symptoms related to the target disease.
\item Search the second-hop neighborhood for drugs used to treat those related diseases and symptoms.
\end{itemize}

\begin{figure}[t]
    \centering
    \includegraphics[width = \linewidth]{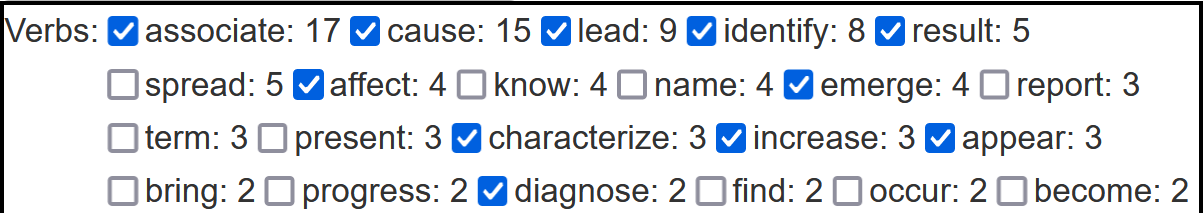}
    \caption{Verb Modifiers between \textit{COVID-19} and \textit{Disease} or \textit{Symptom}. ``Correlation'' related modifiers are checked.}
    \label{fig:dis_verbs}
    \vspace{-2mm}
\end{figure}

\begin{figure}[t]
    \centering
    \includegraphics[width = \linewidth]{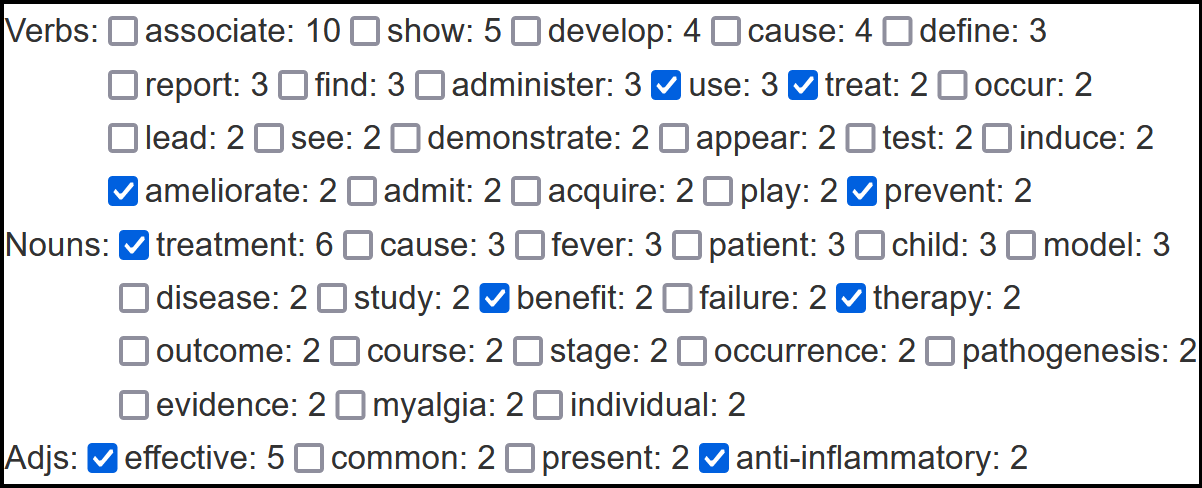}
    \caption{Modifiers between COVID-19 related \textit{Disease} or \textit{Symptom} and \textit{Chemicals} or \textit{Drugs}. ``Treatment'' related modifiers are checked.}
    \label{fig:che_all}
    \vspace{-4mm}
\end{figure}

\begin{table}[tp!]
\small
\centering
    \begin{tabular}{p{0.92\linewidth}}
        \hline
        Candidate drugs\\
         \hline
         nitric oxide, lamb preparation, beta-Lactams, Leukotriene B4, sphingosine 1-phosphate, amoxicillin, Macrolide Antibiotics, Macrolides, beta-Lactams, rifampin, Hydroxymethylglutaryl-CoA Reductase Inhibitors, methylprednisolone, trivalent influenza vaccine, Fibrates, lipid modifying drugs, plain, Corticosteroid ophthalmologic and otologic preparations, metformin, inhibitors, Corticosteroid otologicals, Bilirubin, Fibrates, nitazoxanide, atorvastatin, Artemisinins, antagonists \\
 \\
         \hline
    \end{tabular}
    \caption{Collected candidate drugs for COVID-19 treatment.}
    \label{tab:collected_drugs}
    \vspace{-2mm}
\end{table}

Suppose a researcher wants to discover the candidate drugs for COVID-19. By setting \textit{COVID-19} as the starting node and the \textit{Diseases and Symptoms} entity type as the first-hop neighbors, the system retrieved a set of disease or symptom entities and the frequent modifiers. We select several verb modifiers that might indicate a ``correlation'' relationship between the entities and COVID-19. Then, we update the first-hop neighbors to 10 of these ``correlated'' entities and set 5 \textit{Pharmacologic substance} related entity types as the second-hop neighbors. The retrieved two-hop graph can be seen in Figure \ref{fig:illustration}. Similarly, we select several modifiers that might indicate a ``treatment'' relation to find candidate drugs. The selected modifiers are shown in Figures \ref{fig:dis_verbs} and \ref{fig:che_all} and the collected candidate drugs are listed in Table \ref{tab:collected_drugs}.

\begin{table*}[tp!]
\small
\begin{center}
\begin{tabular}{p{0.2\linewidth}|p{0.35\linewidth}|p{0.35\linewidth}}
\toprule
 & (COVID-19, Pneumonia) & (Pneumonia, Vaccines) \\
\midrule 
{Extracted relation descriptions} &  \textbf{Coronavirus disease 2019} (COVID-19) is a novel type of highly contagious \textbf{pneumonia} caused by the severe acute respiratory syndrome coronavirus 2 (SARS-CoV-2).
 & Despite the availability of safe and effective antibiotics and \textbf{vaccines} for treatment and prevention, \textbf{pneumonia} is a leading cause of death worldwide and the leading infectious disease killer. \\
 & Conversely, SARS-CoV, MERS-CoV, and \textbf{COVID-19} may initially present asymptomatically, but can progress to \textbf{pneumonia}, shortness of breath, renal insufficiency and, in some cases, death.
 & Despite advances in managerial practices, \textbf{vaccines}, and clinical therapies, \textbf{pneumonia} remains a widespread problem and methods to enhance host resistance to pathogen colonization and pneumonia are needed.\\
 \hline
1-hop relation summary & \textbf{COVID-19} is a highly contagious \textbf{pneumonia} caused by the severe acute respiratory syndrome coronavirus 2 (SARS-CoV-2). & Despite the availability of safe and effective antibiotics and \textbf{vaccines} for treatment and prevention, \textbf{pneumonia} remains a major cause of death worldwide. \\
\hline
2-hop relation summary (COVID-19, Vaccines) & \multicolumn{2}{|p{0.73\linewidth}}{\textbf{COVID-19} is a major cause of death worldwide, despite the availability of safe and effective antibiotics and \textbf{vaccines} for treatment and prevention of pneumonia.}\\
\bottomrule
\end{tabular}
\vspace{-1mm}
\caption{Example of relation description extracted or generated by the relation synthesis model.}
\vspace{-3mm}
\label{table:extract_generated}
\end{center}
\end{table*}

\subsection{Case Study 2: Literature Curation}

\begin{figure}[t]
    \centering
    \includegraphics[width = \linewidth]{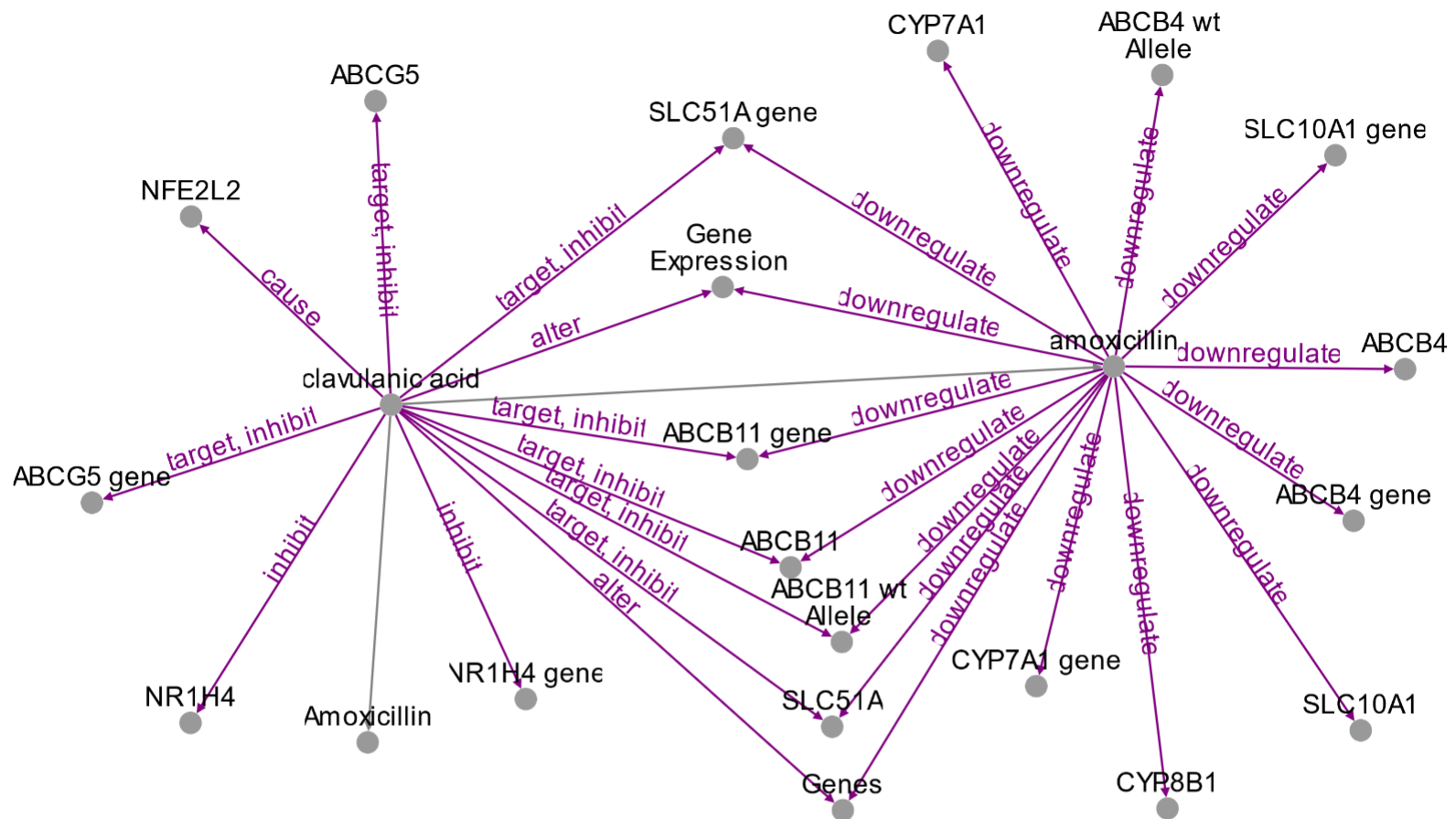}
    \caption{The local graph built from the passage in PMID: 34767876 with \textit{Chemicals} and \textit{Gene} entities.}
    \label{fig:local_deer}
    \vspace{-4mm}
\end{figure}

Literature curation \cite{Wiegers2009TextCTD} is the process of identifying documents relevant to a task or topic and locating and annotating the content of interest in these documents. The latter requires the curator to read through the whole document, which can be tedious and time-consuming. Our system provides an interface where users can run the pipeline in Figure \ref{fig:pipeline} to build a DEER on any article indexed in PubMed and use this graph to locate the information relevant to the curation target.

Suppose a curation task is to collect Drug-Target interactions and a curator is assigned a relevant article. The curator first submits the article's PubMed reference number (PMID) or PMC Identifier (PMCID) to the interface. The interface will return a DEER built from the article and a list of entity types appear in the DEER. Then the curator can read the entity types and select the types that are relevant to the task, for instance, \textit{Chemicals} and \textit{Gene}, and refresh the DEER with only entities of the selected types remaining. An example DEER built from PMID 34767876 with the above three types of entities remaining and some modifiers on the edges is shown in Figure \ref{fig:local_deer}. From the graph, the curator can easily see the possible Drug-Target interactions in the passage and click the edges to verify the information. The content of PMID 34767876 is placed in Appendix \ref{sec:pmid}.

\section{Related Work}

\paragraph{Exploration System}
Exploration systems aim to help users learn the content in the data sources through simple queries \cite{Wang2021TextCOVID-19}. Some systems are designed to retrieve sentence-level text pieces for a specific need. \citet{wang-etal-2020-evidenceminer} retrieve textual evidence that semantically matches the queried statement. \citet{Lahav2022ADirections} build a set of scientific challenges and directions extracted from a corpus and retrieves challenges and directions through entity co-occurrence. \citet{Taub-Tabib2020InteractiveCorpora} develop a lightweight query language to retrieve sentences that syntactically match an example sentence. In contrast, our system collects relational sentences into a graph structure and displays the retrieved sentences in a graphic view that shows the connection between the text pieces, which is not seen in previous works.

\paragraph{Literature-Based Discovery}
Literature-based discovery (LBD) tools aim to discover unknown knowledge and generate new hypotheses by connecting current knowledge scattered in different literature together \cite{Swanson2008}, which is commonly used in biomedical tasks like drug repurposing and interaction prediction. Early LBD tools \cite{swanson1986fish, smalheiser1996indomethacin} require manual effort in organizing information from the passages. Recent studies \cite{PU2023104464} approach LBD as a link prediction task over knowledge bases, where new knowledge is discovered as predicting new links between concepts. Our system is more like the early LBD tools. With the high-quality relational sentences and the advance of LLM in language understanding, our system greatly alleviates the user's reading workload and allows a rough verification of the generated hypothesis based on the retrieved sentences.

\section{Conclusion}

In this work, we developed an exploration system in the biomedical domain that operates on a COVID-related corpus facilitating efficient retrieval of relational knowledge and enabling tasks such as drug repurposing and literature curation. We demonstrate the advantages of managing a raw text corpus in a descriptive knowledge graph, including streamlined management, support for multi-hop reasoning across sentences from various articles, and comprehensive visualization of entity connections in the domain. Additionally, we equipped users with a modifier filtering module and a relation synthesis model that offer an overview of the relations on the edge before reading. In future work, we aim to enhance the accuracy and reliability of the relation descriptions generated for user reference.

\section*{Limitations}
Our system currently only support at most 2-hop query, since the number of entities in the graph will grow exponentially as the path gets longer, which will cause difficulty in reading the graph and reasoning along the path. This hinders our system from studies of more complex network like biochemical pathways, which involves several steps of reaction, and limits the possible knowledge that could be discovered by the system.

The relation synthesis model is trained to generate a single relation description, which is sometimes incapable to cover all the necessary information about the relationship of the target entities. ChatGPT could generate a short paragraph with more details included, but it is more costly than running a local fine-tuned model.

\section*{Acknowledgements}

This material is based upon work supported by the National Science Foundation IIS 16-19302 and IIS 16-33755, Zhejiang University ZJU Research 083650, IBM-Illinois Center for Cognitive Computing Systems Research (C3SR) and IBM-Illinois Discovery Accelerator Institute (IIDAI), grants from eBay and Microsoft Azure, UIUC OVCR CCIL Planning Grant 434S34, UIUC CSBS Small Grant 434C8U, and UIUC New Frontiers Initiative. Any opinions, findings, conclusions, or recommendations expressed in this publication are those of the author(s) and do not necessarily reflect the views of the funding agencies.

\bibliography{anthology,custom}

\begin{thebibliography}{29}
\expandafter\ifx\csname natexlab\endcsname\relax\def\natexlab#1{#1}\fi

\bibitem[{Anil et~al.(2023)Anil, Dai, Firat, Johnson, Lepikhin, Passos, Shakeri, Taropa, Bailey, Chen et~al.}]{anil2023palm}
Rohan Anil, Andrew~M Dai, Orhan Firat, Melvin Johnson, Dmitry Lepikhin, Alexandre Passos, Siamak Shakeri, Emanuel Taropa, Paige Bailey, Zhifeng Chen, et~al. 2023.
\newblock Palm 2 technical report.
\newblock \emph{arXiv preprint arXiv:2305.10403}.

\bibitem[{Bubeck et~al.(2023)Bubeck, Chandrasekaran, Eldan, Gehrke, Horvitz, Kamar, Lee, Lee, Li, Lundberg et~al.}]{bubeck2023sparks}
S{\'e}bastien Bubeck, Varun Chandrasekaran, Ronen Eldan, Johannes Gehrke, Eric Horvitz, Ece Kamar, Peter Lee, Yin~Tat Lee, Yuanzhi Li, Scott Lundberg, et~al. 2023.
\newblock Sparks of artificial general intelligence: Early experiments with gpt-4.
\newblock \emph{arXiv preprint arXiv:2303.12712}.

\bibitem[{Chung et~al.()Chung, Zhou, Pang, Tao, and Zhang}]{ChungBioKDE:Platform}
Meng-Han Chung, Jun Zhou, Xiaodong Pang, Yuchuan Tao, and Jinfeng Zhang.
\newblock \href {https://biokde.com/KG/KG000367} {{BioKDE: a Deep Learning Powered Search Engine and Biomedical Knowledge Discovery Platform}}.

\bibitem[{Handler and O’Connor(2018)}]{handler2018relational}
Abram Handler and Brendan O’Connor. 2018.
\newblock Relational summarization for corpus analysis.
\newblock In \emph{Proceedings of the 2018 Conference of the North American Chapter of the Association for Computational Linguistics: Human Language Technologies, Volume 1 (Long Papers)}, pages 1760--1769.

\bibitem[{Huang et~al.(2022{\natexlab{a}})Huang, Chang, Xiong, and Hwu}]{huang2022open}
Jie Huang, Kevin Chang, Jinjun Xiong, and Wen-mei Hwu. 2022{\natexlab{a}}.
\newblock \href {https://doi.org/10.18653/v1/2022.findings-acl.26} {Open relation modeling: Learning to define relations between entities}.
\newblock In \emph{Findings of the Association for Computational Linguistics: ACL 2022}, pages 297--308, Dublin, Ireland. Association for Computational Linguistics.

\bibitem[{Huang and Chang(2022)}]{huang2022ver}
Jie Huang and Kevin Chen-Chuan Chang. 2022.
\newblock \href {https://arxiv.org/abs/2211.11093} {Ver: Learning natural language representations for verbalizing entities and relations}.
\newblock \emph{ArXiv preprint}, abs/2211.11093.

\bibitem[{Huang and Chang(2023)}]{huang-chang-2023-towards}
Jie Huang and Kevin Chen-Chuan Chang. 2023.
\newblock \href {https://doi.org/10.18653/v1/2023.findings-acl.67} {Towards reasoning in large language models: A survey}.
\newblock In \emph{Findings of the Association for Computational Linguistics: ACL 2023}, pages 1049--1065, Toronto, Canada. Association for Computational Linguistics.

\bibitem[{Huang et~al.(2023)Huang, Gao, Li, Yang, Song, Zhang, Zhu, Jiang, Chang, and Yin}]{huang2023ccgen}
Jie Huang, Yifan Gao, Zheng Li, Jingfeng Yang, Yangqiu Song, Chao Zhang, Zining Zhu, Haoming Jiang, Kevin Chen-Chuan Chang, and Bing Yin. 2023.
\newblock \href {https://arxiv.org/abs/2305.11480} {Ccgen: Explainable complementary concept generation in e-commerce}.
\newblock \emph{arXiv preprint arXiv:2305.11480}.

\bibitem[{Huang et~al.(2022{\natexlab{b}})Huang, Zhu, Chang, Xiong, and Hwu}]{huang-etal-2022-deer}
Jie Huang, Kerui Zhu, Kevin Chen-Chuan Chang, Jinjun Xiong, and Wen-mei Hwu. 2022{\natexlab{b}}.
\newblock \href {https://aclanthology.org/2022.emnlp-main.448} {{DEER}: Descriptive knowledge graph for explaining entity relationships}.
\newblock In \emph{Proceedings of the 2022 Conference on Empirical Methods in Natural Language Processing}, pages 6686--6698, Abu Dhabi, United Arab Emirates. Association for Computational Linguistics.

\bibitem[{Izacard and Grave(2020)}]{Izacard2020LeveragingAnswering}
Gautier Izacard and Edouard Grave. 2020.
\newblock \href {https://doi.org/10.18653/v1/2021.eacl-main.74} {{Leveraging Passage Retrieval with Generative Models for Open Domain Question Answering}}.
\newblock \emph{EACL 2021 - 16th Conference of the European Chapter of the Association for Computational Linguistics, Proceedings of the Conference}, pages 874--880.

\bibitem[{Lahav et~al.(2022)Lahav, Falcon, Kuehl, Johnson, Parasa, Shomron, Chau, Yang, Horvitz, Weld, and Hope}]{Lahav2022ADirections}
Dan Lahav, Jon~Saad Falcon, Bailey Kuehl, Sophie Johnson, Sravanthi Parasa, Noam Shomron, Duen~Horng Chau, Diyi Yang, Eric Horvitz, Daniel~S. Weld, and Tom Hope. 2022.
\newblock \href {https://doi.org/10.1609/AAAI.V36I11.21456} {{A Search Engine for Discovery of Scientific Challenges and Directions}}.
\newblock \emph{Proceedings of the AAAI Conference on Artificial Intelligence}, 36(11):11982--11990.

\bibitem[{Levy et~al.(2021)Levy, Mo, Xiong, and Wang}]{Levy2021Open-DomainDomains}
Sharon Levy, Kevin Mo, Wenhan Xiong, and William~Yang Wang. 2021.
\newblock \href {https://doi.org/10.18653/v1/2021.emnlp-demo.30} {{Open-Domain Question-Answering for COVID-19 and Other Emergent Domains}}.
\newblock \emph{EMNLP 2021 - 2021 Conference on Empirical Methods in Natural Language Processing: System Demonstrations}, pages 259--266.

\bibitem[{Liu et~al.(2023)Liu, Huang, Zhu, and Chang}]{liu-etal-2023-dimongen}
Chenzhengyi Liu, Jie Huang, Kerui Zhu, and Kevin Chen-Chuan Chang. 2023.
\newblock \href {https://doi.org/10.18653/v1/2023.acl-long.260} {{D}imon{G}en: Diversified generative commonsense reasoning for explaining concept relationships}.
\newblock In \emph{Proceedings of the 61st Annual Meeting of the Association for Computational Linguistics (Volume 1: Long Papers)}, pages 4719--4731, Toronto, Canada. Association for Computational Linguistics.

\bibitem[{Neumann et~al.(2019)Neumann, King, Beltagy, and Ammar}]{Neumann2019ScispaCy:Processing}
Mark Neumann, Daniel King, Iz~Beltagy, and Waleed Ammar. 2019.
\newblock \href {https://doi.org/10.18653/v1/w19-5034} {{ScispaCy: Fast and robust models for biomedical natural language processing}}.
\newblock In \emph{BioNLP 2019 - SIGBioMed Workshop on Biomedical Natural Language Processing, Proceedings of the 18th BioNLP Workshop and Shared Task}, pages 319--327. Association for Computational Linguistics (ACL).

\bibitem[{OpenAI(2022)}]{openai2022chatgpt}
OpenAI. 2022.
\newblock Chatgpt: Optimizing language models for dialogue.
\newblock \emph{OpenAI}.

\bibitem[{OpenAI(2023)}]{openai2023gpt4}
OpenAI. 2023.
\newblock \href {http://arxiv.org/abs/2303.08774} {Gpt-4 technical report}.

\bibitem[{Pu et~al.(2023)Pu, Beck, and Verspoor}]{PU2023104464}
Yiyuan Pu, Daniel Beck, and Karin Verspoor. 2023.
\newblock \href {https://doi.org/https://doi.org/10.1016/j.jbi.2023.104464} {Graph embedding-based link prediction for literature-based discovery in alzheimer’s disease}.
\newblock \emph{Journal of Biomedical Informatics}, page 104464.

\bibitem[{Qin et~al.(2023)Qin, Zhang, Zhang, Chen, Yasunaga, and Yang}]{qin2023chatgpt}
Chengwei Qin, Aston Zhang, Zhuosheng Zhang, Jiaao Chen, Michihiro Yasunaga, and Diyi Yang. 2023.
\newblock Is chatgpt a general-purpose natural language processing task solver?
\newblock \emph{arXiv preprint arXiv:2302.06476}.

\bibitem[{Shen et~al.(2018)Shen, Xiao, He, Shang, Sinha, and Han}]{Shen2018EntityApproach}
Jiaming Shen, Jinfeng Xiao, Xinwei He, Jingbo Shang, Saurabh Sinha, and Jiawei Han. 2018.
\newblock \href {https://doi.org/10.1145/3209978.3210055} {{Entity Set Search of Scientific Literature: An Unsupervised Ranking Approach}}.
\newblock \emph{41st International ACM SIGIR Conference on Research and Development in Information Retrieval, SIGIR 2018}, pages 565--574.

\bibitem[{Smalheiser and Swanson(1996)}]{smalheiser1996indomethacin}
Neil~R Smalheiser and Don~R Swanson. 1996.
\newblock Indomethacin and alzheimer's disease.
\newblock \emph{Neurology}, 46(2):583--583.

\bibitem[{Swanson(2008)}]{Swanson2008}
D.~R. Swanson. 2008.
\newblock \href {https://doi.org/10.1007/978-3-540-68690-3_1} {\emph{Literature-Based Discovery? The Very Idea}}, pages 3--11. Springer Berlin Heidelberg, Berlin, Heidelberg.

\bibitem[{Swanson(1986)}]{swanson1986fish}
Don~R Swanson. 1986.
\newblock Fish oil, raynaud's syndrome, and undiscovered public knowledge.
\newblock \emph{Perspectives in biology and medicine}, 30(1):7--18.

\bibitem[{Taub-Tabib et~al.(2020)Taub-Tabib, Shlain, Sadde, Lahav, Eyal, Cohen, and Goldberg}]{Taub-Tabib2020InteractiveCorpora}
Hillel Taub-Tabib, Micah Shlain, Shoval Sadde, Dan Lahav, Matan Eyal, Yaara Cohen, and Yoav Goldberg. 2020.
\newblock \href {https://doi.org/10.48550/arxiv.2006.04148} {{Interactive Extractive Search over Biomedical Corpora}}.
\newblock \emph{BioNLP}, pages 28--37.

\bibitem[{Voorhees et~al.(2021)Voorhees, Alam, Bedrick, Demner-Fushman, Hersh, Lo, Roberts, Soboroff, and Wang}]{10.1145/3451964.3451965}
Ellen Voorhees, Tasmeer Alam, Steven Bedrick, Dina Demner-Fushman, William~R. Hersh, Kyle Lo, Kirk Roberts, Ian Soboroff, and Lucy~Lu Wang. 2021.
\newblock \href {https://doi.org/10.1145/3451964.3451965} {Trec-covid: Constructing a pandemic information retrieval test collection}.
\newblock \emph{SIGIR Forum}, 54(1).

\bibitem[{Wang and Lo(2021)}]{Wang2021TextCOVID-19}
Lucy~Lu Wang and Kyle Lo. 2021.
\newblock \href {https://doi.org/10.1093/BIB/BBAA296} {{Text mining approaches for dealing with the rapidly expanding literature on COVID-19}}.
\newblock \emph{Briefings in Bioinformatics}, 22(2):781--799.

\bibitem[{Wang et~al.(2020{\natexlab{a}})Wang, Lo, Chandrasekhar, Reas, Yang, Burdick, Eide, Funk, Katsis, Kinney, Li, Liu, Merrill, Mooney, Murdick, Rishi, Sheehan, Shen, Stilson, Wade, Wang, Wang, Wilhelm, Xie, Raymond, Weld, Etzioni, and Kohlmeier}]{wang-etal-2020-cord}
Lucy~Lu Wang, Kyle Lo, Yoganand Chandrasekhar, Russell Reas, Jiangjiang Yang, Doug Burdick, Darrin Eide, Kathryn Funk, Yannis Katsis, Rodney~Michael Kinney, Yunyao Li, Ziyang Liu, William Merrill, Paul Mooney, Dewey~A. Murdick, Devvret Rishi, Jerry Sheehan, Zhihong Shen, Brandon Stilson, Alex~D. Wade, Kuansan Wang, Nancy Xin~Ru Wang, Christopher Wilhelm, Boya Xie, Douglas~M. Raymond, Daniel~S. Weld, Oren Etzioni, and Sebastian Kohlmeier. 2020{\natexlab{a}}.
\newblock \href {https://aclanthology.org/2020.nlpcovid19-acl.1} {{CORD-19}: The {COVID-19} open research dataset}.
\newblock In \emph{Proceedings of the 1st Workshop on {NLP} for {COVID-19} at {ACL} 2020}, Online. Association for Computational Linguistics.

\bibitem[{Wang et~al.(2021)Wang, Li, Wang, Parulian, Han, Ma, Tu, Lin, Zhang, Liu, Chauhan, Guan, Li, Li, Song, Fung, Ji, Han, Chang, Pustejovsky, Rah, Liem, ELsayed, Palmer, Voss, Schneider, and Onyshkevych}]{wang-etal-2021-covid}
Qingyun Wang, Manling Li, Xuan Wang, Nikolaus Parulian, Guangxing Han, Jiawei Ma, Jingxuan Tu, Ying Lin, Ranran~Haoran Zhang, Weili Liu, Aabhas Chauhan, Yingjun Guan, Bangzheng Li, Ruisong Li, Xiangchen Song, Yi~Fung, Heng Ji, Jiawei Han, Shih-Fu Chang, James Pustejovsky, Jasmine Rah, David Liem, Ahmed ELsayed, Martha Palmer, Clare Voss, Cynthia Schneider, and Boyan Onyshkevych. 2021.
\newblock \href {https://doi.org/10.18653/v1/2021.naacl-demos.8} {{COVID}-19 literature knowledge graph construction and drug repurposing report generation}.
\newblock In \emph{Proceedings of the 2021 Conference of the North American Chapter of the Association for Computational Linguistics: Human Language Technologies: Demonstrations}, pages 66--77, Online. Association for Computational Linguistics.

\bibitem[{Wang et~al.(2020{\natexlab{b}})Wang, Guan, Liu, Chauhan, Jiang, Li, Liem, Sigdel, Caufield, Ping, and Han}]{wang-etal-2020-evidenceminer}
Xuan Wang, Yingjun Guan, Weili Liu, Aabhas Chauhan, Enyi Jiang, Qi~Li, David Liem, Dibakar Sigdel, John Caufield, Peipei Ping, and Jiawei Han. 2020{\natexlab{b}}.
\newblock \href {https://doi.org/10.18653/v1/2020.acl-demos.8} {{EVIDENCEMINER}: Textual evidence discovery for life sciences}.
\newblock In \emph{Proceedings of the 58th Annual Meeting of the Association for Computational Linguistics: System Demonstrations}, pages 56--62, Online. Association for Computational Linguistics.

\bibitem[{Wiegers et~al.(2009)Wiegers, Davis, Cohen, Hirschman, and Mattingly}]{Wiegers2009TextCTD}
Thomas~C. Wiegers, Allan~P. Davis, K.~Bretonnel Cohen, Lynette Hirschman, and Carolyn~J. Mattingly. 2009.
\newblock \href {https://doi.org/10.1186/1471-2105-10-326/FIGURES/5} {{Text mining and manual curation of chemical-gene-disease networks for the Comparative Toxicogenomics Database (CTD)}}.
\newblock \emph{BMC Bioinformatics}, 10(1):326.

\end{thebibliography}
\bibliographystyle{acl_natbib}

\clearpage

\appendix

\section{Content of PMID 34767876}
\label{sec:pmid}

\noindent\fbox{
    \parbox{\linewidth}{
    \small
        Molecular mechanisms of hepatotoxic cholestasis by clavulanic acid: Role of NRF2 and FXR pathways.\\

        Treatment of beta-lactamase positive bacterial infections with a combination of amoxicillin (AMOX) and clavulanic acid (CLAV) causes idiosyncratic drug-induced liver injury (iDILI) in a relevant number of patients, often with features of intrahepatic cholestasis. This study aims to determine serum bile acid (BA) levels in amoxicillin/clavulanate (A+C)-iDILI patients and to investigate the mechanism of cholestasis by A+C in human in vitro hepatic models. In six A+C-iDILI patients, significant elevations of serum primary conjugated BA definitely demonstrated A+C-induced cholestasis. In cultured human Upcyte hepatocytes and HepG2 cells, CLAV was more cytotoxic than AMOX, and, at subcytotoxic concentrations, it altered the expression of more than 1,300 genes. CLAV, but not AMOX, downregulated the expression of key genes for BA transport (BSEP, NTCP, OSTalpha and MDR2) and synthesis (CYP7A1 and CYP8B1). CLAV also caused early oxidative stress, with reduced GSH/GSSG ratio, along with induction of antioxidant nuclear factor erythroid 2-related factor 2 (NRF2) target genes. Activation of NRF2 by sulforaphane also resulted in downregulation of NTCP, OSTalpha, ABCG5, CYP7A1 and CYP8B1. CLAV also inhibited the BA-sensor farnesoid X receptor (FXR), in agreement with the downregulation of FXR targets BSEP, OSTalpha and ABCG5. We conclude that CLAV, the culprit molecule in A+C, downregulates several key biliary transporters by modulating NRF2 and FXR signaling, thus likely promoting intrahepatic cholestasis. On top of that, increased ROS production and GSH depletion may aggravate the cholestatic injury by A+C.
    }
}

\section{Relation Synthesis Model Fine-tuning}
\label{sec:fine-tune}

As the CORD-19 dataset we use for demonstration is not large enough to train a relation synthesis model, we collected a training, validation, and test dataset from a subset of articles randomly selected from PubMed. All the target sentences have an RDS score greater than 0.75 and all the input sentences have an RDS score greater than 0.7. This resulted in a total of 615,561, 12,824, and 12,825 data in the training, validation, and test dataset respectively, which is comparable in size to the one used in DEER. We trained the model for 20 epochs. Other settings are the same as \citet{huang-etal-2022-deer}. In Section \ref{sec:evaluation}, we discuss our manual evaluation of the quality of generation.

\section{Relation Synthesis with ChatGPT}
\label{sec:prompt}

The input to ChatGPT consists of a prompt and an overall relation context.
\begin{itemize}[leftmargin=*, nolistsep, nolistsep, topsep=1mm]
\setlength{\itemsep}{1mm}
    \item The prompt is ``Given the context below, describe the relation between [target\_entity\_1] and [target\_entity\_2] in one sentence.'' with [target\_entity\_1] and [target\_entity\_2] replaced by the target entity pair.
    \item The overall relation context is formed by concatenating the relation context on each edge with a new line character. The relation context on each edge starts with a description ``Relation between [head\_entity] and [tail\_entity]:\\n'' with [head\_entity] and [tail\_entity] replaced by the head and tail entity of the edge and is followed by the top 5 sentences on the edge sorted by the RDS score and concatenated by the new line character.
\end{itemize}

Using the sentences in Table \ref{table:extract_generated} as an example, the input for ChatGPT is

\noindent\fbox{
    \parbox{\linewidth}{
    \small
        Given the context below, describe the relation between COVID-19 and Vaccines in one sentence.\\

        Relation between COVID-19 and Pneumonia:\\
        Coronavirus disease 2019 (COVID-19) is a novel type of highly contagious pneumonia caused by the severe acute respiratory syndrome coronavirus 2 (SARS-CoV-2).\\
        Conversely, SARS-CoV, MERS-CoV, and COVID-19 may initially present asymptomatically, but can progress to pneumonia, shortness of breath, renal insufficiency and, in some cases, death.\\

        Relation between Pneumonia and Vaccines:\\
        Despite the availability of safe and effective antibiotics and vaccines for treatment and prevention, pneumonia is a leading cause of death worldwide and the leading infectious disease killer.\\
        Despite advances in managerial practices, vaccines, and clinical therapies, pneumonia remains a widespread problem and methods to enhance host resistance to pathogen colonization and pneumonia are needed.
    }
}

\end{document}